\documentclass[sigconf]{acmart}

\usepackage{booktabs} 
\usepackage[utf8]{inputenc} 
\usepackage[T1]{fontenc}    
\usepackage{amsfonts}       
\usepackage{nicefrac}       
\usepackage{microtype}      
\usepackage{enumitem}
\usepackage{placeins}
\usepackage{dsfont}

\usepackage{amsmath}
\usepackage{color}
\usepackage{amssymb}

\usepackage{color,soul}
\usepackage{mathtools}

\usepackage{graphicx}
\graphicspath{ {images/} }

\usepackage{caption}
\usepackage{subcaption}

\usepackage[colorinlistoftodos]{todonotes}

\usepackage{bm}

\setcopyright{rightsretained}

\acmDOI{10.475/123_4}

\acmISBN{123-4567-24-567/08/06}

\acmConference[KDD'19]{ACM KDD Conference}{August 2019}{Anchorage, AK}
\acmYear{2019}
\copyrightyear{2019}

\acmArticle{4}
\acmPrice{15.00}

\begin{document}
\title{Spatial sensitivity analysis for urban land use prediction with physics-constrained conditional generative adversarial networks}

\author{Adrian Albert}
\authornote{Corresponding author. This research was performed while the author was with the Massachusetts Institute of Technology (MIT) in Cambridge, MA.}
\affiliation{%
  \institution{Lawrence Berkeley National Laboratory}
  \streetaddress{1 Cyclotron Rd}
  \city{Berkeley}
  \state{CA}
  \postcode{94720}
}

\email{aalbert@lbl.gov}

\author{Emanuele Strano}

\affiliation{%
  \institution{MindEarth}
  \streetaddress{}
  \city{Biel/Bienne}
  \country{Switzerland}
}
\email{emanuele.strano@mindearth.org}

\author{Jasleen Kaur}
\affiliation{%
  \institution{Phillips Lighting Research}
  \streetaddress{2 Canal St}
  \city{Cambridge}
  \state{MA}
}
\email{jasleen.kaur1@philips.com}

\author{Marta C. Gonzalez}
\affiliation{%
  \institution{University of California, Berkeley}
  \city{Berkeley}
  \state{CA}
  \postcode{94720}
}
\email{martag@berkeley.edu}

\renewcommand{\shortauthors}{A. Albert et al.}

\begin{abstract}
Accurately forecasting urban development and its environmental and climate impacts critically depends on realistic models of the spatial structure of the built environment, and of its dependence on key factors such as population and economic development. Scenario simulation and sensitivity analysis, i.e., predicting how changes in underlying factors at a given location affect urbanization outcomes at other locations, is currently not achievable at a large scale with traditional urban growth models, which are either too simplistic, or depend on detailed locally-collected socioeconomic data that is not available in most places. Here we develop a framework to estimate, purely from globally-available remote-sensing data and without parametric assumptions, the spatial sensitivity of the (\textit{static}) rate of change of urban sprawl to key macroeconomic development indicators. We formulate this spatial regression problem as an image-to-image translation task using conditional generative adversarial networks (GANs), where the gradients necessary for comparative static analysis are provided by the backpropagation algorithm used to train the model. This framework allows to naturally incorporate physical constraints, e.g., the inability to build over water bodies. To validate the spatial structure of model-generated built environment distributions, we use  spatial statistics commonly used in urban form analysis. We apply our method to a novel dataset comprising of layers on the built environment, nightlighs measurements (a proxy for economic development and energy use), and population density for the world's most populous 15,000 cities. 
\end{abstract}

%
%



\maketitle

\section{Introduction}
\label{sec:introduction}

The overwhelming impact of the built environment on a wide range of global issues such as climate change, energy use, and economic development and activity, is hard to overstate. Buildings consume 60\% of the world's energy, more than 50\% of the world's population lives in cities, which are responsible for 70\% of GHG emissions and 80\% of economic output globally \cite{zed-paper}. As such, realistic models of the evolution of the spatial distribution of the urban built environment, and its dependence on key socio-economic factors, are of high relevance for urban planning, energy management, and infrastructure investments. However, the tools to study urban form at a global scale have to date been largely based on simplified, bottom-up models \cite{rybski_2013} that fail to capture the complexity observed in real data and to produce realistic predictions of the spatio-temporal dynamics of land use. Further key limitations of traditional urban development models are \textit{i)} their inability to effectively leverage the vast, ever-increasing amount of observational data on urban built areas, and \textit{ii)} their dependence on detailed, local socio-economic data that are not available in the vast majority of cities.

Specific examples of applications where urbanization forecasts are needed, but on-the-ground data is not available or is difficult to obtain that we have identified with partners such as the World Bank include \textit{i)} the management and inclusion of large refugees camps within the urbanization plans of cities in Africa; \textit{ii)} the economic evaluation and forecasting of necessary future urban investments in central Africa; and \textit{iii)} informing carbon emission, energy use, and water consumption forecasting for Asian and African regions. 

In this paper, our goal is to build a machine learning simulation model of the spatial structure of the urban built land use that \textit{i)} produces predictions (in a comparative statics sense, of the next time step) that are realistic, both qualitatively, and quantitatively, \textit{ii)} can naturally incorporate physical constraints, and \textit{iii)} can easily be used to perform sensitivity analysis, at a large scale, of key underlying socio-economic factors affecting urbanization such as population density and (proxies for) economic activity. The model would generate simulations, or``synthetic" cities, in a controlled way, to be used in scenario analyses and urbanization forecasts.

For this, we developed our earlier work \cite{albert_igarss_2018} and propose a conditional generative adversarial network (GAN) \cite{gan_paper_2014} model that formulates our task - a spatial regression - as an image-to-image translation problem, inspired by \cite{pix2pix_2016}. We develop our GAN model to predict the spatial distribution of built-up urbanized land starting from either \textit{i)} random noise and input maps encoding physical constraints such as water areas (where nothing can be built), and, in addition, \textit{ii)} maps of population density and luminosity levels (a rough proxy for economic development \cite{nightlights_wealth_2017}). This formulation allows to naturally incorporate constraints such as the inability to develop over water bodies by adding appropriate penalty terms in the GAN loss function. To measure the quality of generated maps, we compare real and generated built-up patterns maps using domain-inspired spatial statistics routinely used in urban analysis, economic geography,  and ``urban science" literature \cite{makse1995modelling}, such as built-up area density and fractal dimension of built-up patterns. We note that the spatial sensitivity of urban built land use density to underlying factors such as population density and economic activity are practically spatial derivatives obtained as a by-product of the backpropagation algorithm used for training. 

To train our model, we use a novel, global-scale dataset of remote-sensing data products as detailed in \cite{albert_kdd_urbcomp_2018}. This dataset includes a unique data layer on built-up areas at global scale obtained from the German Aerospace Center (DLR) \cite{esch_sar_guf_2013}. Further, the dataset incorporates remote-sensing data on population density from the LandScan project \cite{landscan-data}, and on nightlights as proxy for local economic activity, e.g.,  \cite{henderson_nighlights_gdp_2011}, and of energy use \cite{seto_nightlights_energy_2017} from NASA's VIIRS mission \cite{viirs-data} (see Section \ref{sec:experimental}).

The paper is organized as follows: in Section \ref{sec:literature} we review related literature and current urban modeling approaches. Section \ref{sec:model} describes the GAN model. Section \ref{sec:experimental} discusses our experimental setup and data. We discuss results in Section \ref{sec:results}, and conclude in Section \ref{sec:conclusions}. All the code and data will be made available shortly. 
\section{Related literature}
\label{sec:literature}

\subsection{Modeling urbanization and built land use}
Traditional spatial explicit urban evolutionary models can be categorized in two main classes: agent based modeling (ABM) and complex systems modeling (CSM) \cite{BattyBook}. Both of them, as reported repeatedly in the urban growth modeling literature, are not effective tools for decision making \cite{landis2011urban}. The reason of their failure can be summarized as follows. 

The ABM approach  is based on a probabilistic model that assigns to a non-urban area the probability to become urban after a period of time. This probability is calibrated with a wide and detailed variety of spatial co-variants such as population density, price of the land, mobility, population demographic profiles, etc. Collecting spatial co-variants (typically through on-the-ground surveys) is a painstaking, long, difficult and expensive process and most of the time, by the end of the survey process, the area of interest is already changed. Moreover, in the very the areas where cities will growth more in next decades, such as African cities, this kind of spatial and \textit{in-situ} information, critical to inform such models, are minimal or even absent. Thus, the ABM approaches need highly-granular data on urban development that are either very expensive or simply impossible to collect at scale. More agile and adaptable models are necessary to impact decision making processes.

The CSM framework employs a physics-like approach that needs no \textit{in-situ} spatial and economic co-variants apart of urban footprints at a reference point in time. Then it assigns new urbanized areas based on simple assumptions such as bigger cities are more attractive than small ones (gravity law). This approach is much faster and simple than ABM, however, while it reproduces macroscopic urban characteristics such fractality \cite{Batty1994} and scale-free behaviour \cite{Gabaix2004}, it is not able to reproduce realistic urban forms, i.e., the complex spatial patterns of land cover observed in cities. In a way, since CSM searches for universal laws of urbanization, its level of abstractions reproduces urban prototypes that are are too abstract and unrealistic and cannot be used in real-world scenarios.

In short, both types of traditional models (ABM and CSM) are impractical for forecasting real-world urbanization in a low-data regime such as the African continent. As such, there is a pronounced disconnect between the current urban science academic literature and the practical needs of policymakers who have to allocate resources based on urbanization forecasts. Moreover, the forecasting models work at very low resolution (500m to 1km), thus being unable to provide enough granularity to be useful for on-the-ground operational decision making.

\vspace*{-0.5em}
\subsection{GANs for urban science applications}
\vspace*{-0.1em}
Since their introduction in 2014, GANs \cite{gan_paper_2014} have proven to be very effective at addressing long-standing challenges in computer vision (e.g., natural image generation, image super-resolution), speech synthesis, and language translation, among others. Overviews of early techniques used for assessing model quality of different architectural choices are given e.g., in \cite{improved_gans_2016}. GANs have been recently shown to excel at generating data across a variety of disciplines, including, close to our topic of interest, land use modeling \cite{albert_igarss_2018}, remote sensing data processing, and climate modeling \cite{jinlong_gan_2018}. As such, there is now a consistent body of evidence that GANs excel at implicitly sampling from highly complex, analytically-unknown distributions in a great number of contexts.

In previous research \cite{albert_igarss_2018}, we have shown the effectiveness of a unconstrained, unconditional GAN model at generating realistic built land use maps. There, we show that model-generated ``cities'' (that is, maps of built land use) display a high degree of similarity on several statistics used in the urban modeling literature with real cities. The conditional GAN approach we develop in this paper builds upon the work in \cite{albert_igarss_2018}, in addition incorporating (soft) physical constraints such as water areas and the ability to predict land use maps from underlying socio-economic factors. Our model is superior to traditional models in several key aspects. First, it is trained on satellite data products available at a large scale and in most cases globally, making the data collection much faster, objective and scalable than extensive in-situ surveys. Second, it allows an easy, principled way to perform sensitivity analysis and scenario simulation. Third, it casts urbanization modeling as a machine learning problem, which allows us to leverage the tremendous advances in AI models, software frameworks (e.g., PyTorch and Tensorflow), and hardware (GPUs and TPUs). 
\section{Model} 
\label{sec:model}

\subsection{Comparative statics and sensitivity analysis}
Consider the spatial process of urbanization at a given instant in time $x_B$ as a function of several inter-related variables $x_A$, among which population density and the amount of ``economic development" (of which nighttime luminosity is a noisy proxy \cite{henderson_nighlights_gdp_2011}). Assume that the urban built land use change process can be expressed in closed form as $0=F(t,x_B(t), x_A(t))$. For small changes in both $t$ and $x_A$, it is easy to see that the change of $x_B$ (its total derivative) can then be modeled as a linear combination of a time-dependent part (with no spatial variation) and a space-dependent part (with no temporal variation). Here we focus only on the short-term spatial variation at a given time instant $t$. In econometrics, this procedure of analyzing how endogenous factors change in response to changes in dependent variables is termed \textit{comparative statics}, because it is agnostic of the time-path of the change process, and only focuses on before-and-after comparisons. A typical spatial regression (e.g., \cite{spatial_regression_urban_2013}) would model this as:
\begin{equation}
    x_B = G(x_A) + z = x_\text{pop} \gamma_{pop} + x_\text{lum} \gamma_{lum} + z \text{ (linear regression)},
    \label{eq:regression}
\end{equation}
where $x_A=(x_{pop},x_{lum})$, the $pop$ and $lum$ subscripts refer to nighttime luminance and population density, $\gamma$ refers to regression coefficients, i.e., $\gamma = \frac{\partial x_B}{\partial x_A} = \left(\frac{\partial x_B}{\partial x_{pop}}, \frac{\partial x_B}{\partial x_{lum}}\right)$, and $z$ is the noise term.

Typical spatial statistics methods make assumptions about the structure of the noise term $z$, e.g., modeling it as a Gaussian process with a covariance that depends on features computed locally \cite{datta0hierarchical}. Moreover, the optimization typically is formulated via \textit{local} loss terms, which average over statistics computed around a given location. Here, we make no assumptions about the form of $z$ or about how $x_A$ and $x_B$ might interact, only that these (i.e., $G(\cdot)$) can be modeled by a neural network.

\subsection{GANs for image-to-image translation}
\begin{figure}
    \centering
    \hspace*{-1em}
    \includegraphics[scale=0.4]{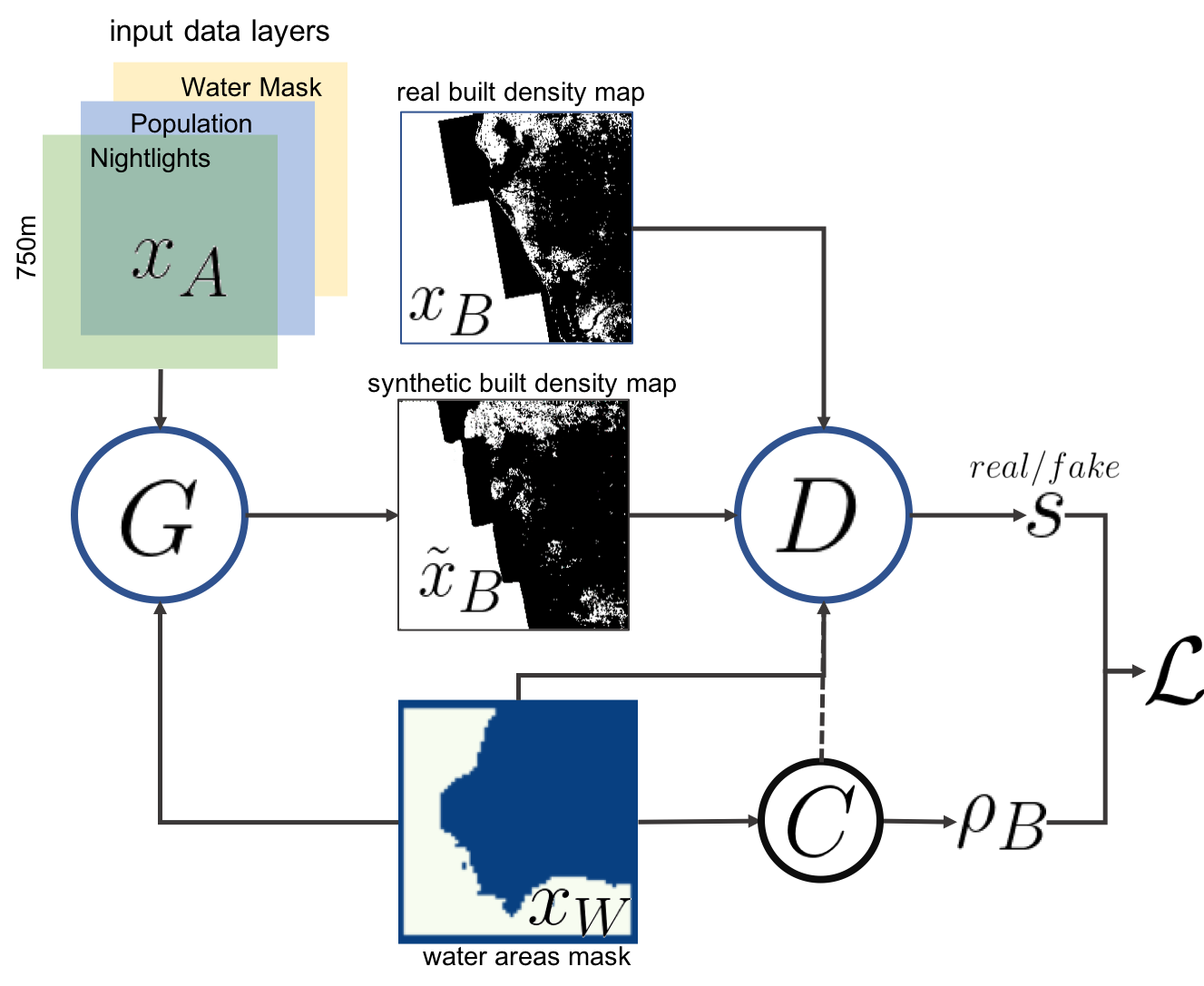}
    \caption{Model architecture for spatial regression with physics-informed conditional GANs. We modify the image-to-image translation architecture in \cite{pix2pix_2016} by adding an additional \textit{constraint-enforcing} model $C$, whose purpose is to guide training towards physically-sound solutions.\vspace*{-0.5em}}
    \label{fig:model_architecture}
\end{figure}%
Generative adversarial networks (GANs) \cite{gan_paper_2014} learn unsupervised representations of input data by training two networks (a \textit{generator} and a \textit{discriminator}) against each other. In the original formulation \cite{gan_paper_2014}, the generator $G$ receives as input a random noise vector $z$, which it transforms in a deterministic way (e.g., by passing it through successive deconvolutional layers if $G$ is a deep CNN) to output a sample $x_\text{fake} = G(z)$. When $G$ is optimal, $x_\text{fake}$ is implicitly sampled from the data distribution that $G$ seeks to emulate. The discriminator $D$ takes an input $x$ (which can be either real, from an empirical dataset, or synthetically generated by $G$), and outputs the source probability $P(s | x) = D(x)$ that $x$ is either sampled from the real distribution ($s=\text{real}$), or produced by $G$ ($s=\text{fake}$). 

The noise input $z$ is unstructured for basic GANs, which makes difficult controlling the generation process for more complex tasks such as simulating urban form with specific properties. In the context of image-to-image translation problems it has been proposed \cite{pix2pix_2016} to replace $z$ and $c$ with structured input (images). The task here is to transform an image from domain $A$ (here, population and luminosity data layers) to domain $B$ (here, maps of built areas density). GAN-based methods that have recently been proposed achieve this via paired samples from the two domains $\{(x^i_A, x^i_B)\}_{i=1}^N$ \cite{pix2pix_2016}. For paired data, the optimization objective function is of the form \cite{pix2pix_2016}:
\begin{align*}
    \mathcal L_{\text{pix2pix}} =& ~ \mathcal L_{cGAN} (G,D) + \lambda \mathcal L_{L_1} (G) (\lambda \text{ is a trade-off parameter})\\
    \mathcal L_{cGAN}(G,D) =& ~ \mathbb E_{(x_A,x_B)\sim p(x_A,x_B)}[\text{log } D(x_A, x_B)] + \nonumber \\
    					   ~& ~ \mathbb E_{x_A\sim p(x_A), z\sim p_z{z}}[\text{log } [1-D(x_A,G(z;x_A))]\\
    \mathcal L_{L_1} = & ~ \mathbb E_{(x_A,x_B)\sim p(x_A,x_B), z\sim p_z{z}} || x_B - G(z;x_A) ||_1.
\end{align*}
$\mathcal L_{cGAN}$ is the typical GAN objective ensuring that the generated images are realistic (i.e., sampled from the real image manifold). The term $\mathcal L_{L_1}$ is a reconstruction error that enforces visual similarity of the generated image $\tilde x_B = G(z; x_A)$ with the real image $x_B$. 

\subsection{Conditional physics-constrained GANs}
The structure of our final model architecture is shown in Figure \ref{fig:model_architecture}. Our first modeling goal is to control the generation process as to simulate realistic scenarios of the effect of changes in the input maps $x_A$ (nightlights, population density) to the output map of built areas $x_B$. The ``pix2pix" architecture \cite{pix2pix_2016} that we build on trains a generator $G$ that can produce realistic renderings of samples from input domain $A$ into domain $B$. As in \cite{pix2pix_2016}, the form of noise $z$ that we allow is via dropout (with $p=0.2$). 

Our second modeling goal is to incorporate physical constraints into GAN model training. For this, we modify the \textit{pix2pix} architecture to include a constraint-enforcing module $C$. This module itself can be a complex model such as a neural network performing a more complicated task. However, here we opt for a simple computation of a error term $\rho_B \equiv \mathds 1 \lbrace \tilde x_B>0 ~ \& ~ x_W>0 \rbrace$ that penalizes the model whenever it generates urbanized land patches over water areas through . Information on water areas is encoded in a water mask $x_{W}$ that is passed both to the generator $G$ as well as the discriminator $D$ and the constraint-enforcing model $C$. This is simple way to impose a soft constraint that does not guarantee the desired solutions by design; however, in practice, we found it is very effective at enforcing the basic physical constraint we are interested in (water areas). Our final optimization objective is:
\begin{align}
    \mathcal L = & ~ L_{\text{pix2pix}} + \mathcal L_{\text{constr}}, \text{ with}\\ 
    \mathcal L_{\text{constr}} = & ~ \alpha \mathbb E_{\tilde x_B\sim p(\tilde x_B)}[\rho_B] \nonumber \\
    					   = & ~ \alpha \mathbb E_{\tilde x_B\sim p(\tilde x_B)}[\mathds 1 \lbrace \tilde x_B>0 ~ \& ~ x_W>0 \rbrace],
\end{align}
where $\alpha$ is a penalty term that we chose to have a large value (here, $\alpha=100$) as to ensure the model has an incentive not to generate urbanized land over water areas. 

\subsection{Conditional GANs as regression models}
In effect, this formulation amounts to a regression of the output built map $\tilde x_B = G(x_A) + z$, with additional regularization provided by $\mathcal L_{cGAN}$ and $\mathcal L_\text{constr}$. The GAN term $\mathcal L_{cGAN}$ forces the generator to output samples from the manifold of empirically-observed built maps $x_B$, which ensures the realism of the regression predictions $\tilde x_B$. The penalty term $\mathcal L_\text{constr}$ ensures compliance with the imposed physical constraint. These two loss terms are \textit{structured}, as they penalize either the whole output map (in the case of $\mathcal L_{cGAN}$) or global geometric properties of the output (in the case of $\mathcal L_\text{constr}$). In addition to these two key points, this framework allows to model critical spatial regression components:

\par \textit{Coefficients and standard errors.} The deterministic input-output transformations are much simpler in standard regressions formulations (e.g., a linear transformation). This allows for straightforward interpretability and statistical tests on the properties of coefficients (gradients of dependent variable with respect to the input covariates). Here the dependence of the output to the input is modeled via a highly-complex neural network. However, this transformation being a composition of simple, differentiable functions, the gradient is readily available as a by-product of the backpropagation algorithm used to train the network. Standard errors on the coefficients may then be estimated via the inverse of the Hessian, which is available readily in certain computational frameworks.
\par \textit{Structure of noise term.} Instead of the typical spatial regression formulation in \ref{eq:regression}, where the structure in the noise term $z$ is explicitly and analytically modelled, our formulation incorporates noise implicitly through dropout operations and leaky ReLU activation functions. Note that if one would wish to impose an explicitly-modelled structure on the noise $z=z(\theta)$, with $\theta$ a vector of parameters, it would be possible to estimate $\theta$ in the same end-to-end process via backpropagation, as long as $z(\cdot)$ is differentiable in $\theta$ and $z$ is part of the computational graph of the model. 

\section{Experimental setup}
\label{sec:experimental}

\subsection{Global remote-sensing urban data layers}
\begin{figure}[h!]
    \centering
    \hspace*{-3em}
    \includegraphics[scale=0.5]{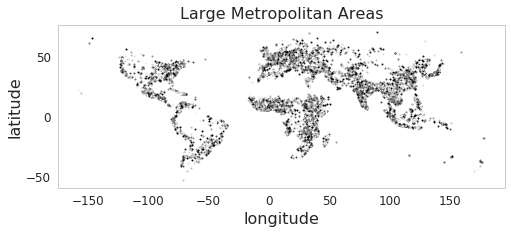}
    \caption{Geographical distribution of the world-wide urban data on urban areas used in this paper.}
    \label{fig:3k_cities_map}
\end{figure}%

\begin{figure}
    \centering
    \hspace*{-2em}
    \includegraphics[scale=0.45]{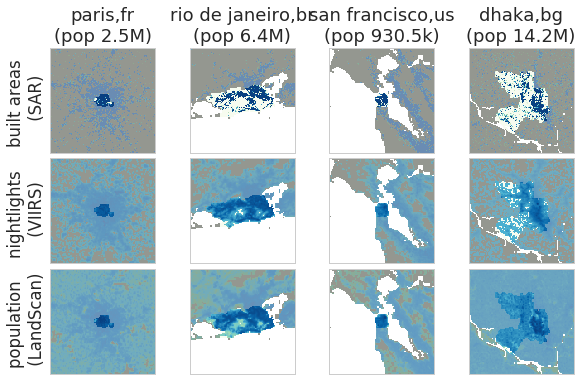}
    \caption{Samples from the \textit{CityNet} dataset \cite{albert_kdd_urbcomp_2018}  (cities on columns, data sources on rows). Example major cities, with water areas in white and city bounds gray-shaded.}
    \label{fig:example_data}
\end{figure}%

\par \textbf{Data layers.} We observe spatial maps $\mathbf{x}^i$ for $i=1,...,N$ cities, with $\mathbf{x}^i \in \mathbb{R}^{W \times W \times S}$, where $S$ is the number of data sources, as well as corresponding binary masks $x^i_W \in \mathbb{R}^{W \times W}$, with $x^i_W(x,y)=1$ if the land from city $i$ at location $(x,y)$ can be developed (here, water areas), and $b_i \in \mathbb{R}^{W \times W}$, representing best-available administrative boundaries for each city $i$, with $b_i(x,y)=1$ if location $(x,y)$ is within the city boundary. $S$ refers to the number of data sources, which are assumed distinct, if not independent, in the sense that each brings additional, interpretable information. Here, we represent a city by $S=3$ data sources: population density $x_{pop}$, nighttime luminosity $x_{lum}$, and building density $x_{bld}$. We thus represent a city by the data layers $\mathbf{x} \equiv [x_{pop}, \ x_{lum}, \ x_{bld}, x_W, b]$.

\par \textbf{The ``CityNet'' world cities dataset.} We used a dataset that was recently introduced in  \cite{albert_kdd_urbcomp_2018} that contains the spatial maps as defined above on world-wide cities with at least $10,000$ inhabitants, i.e., $\sim 32,000$ world-wide. For each of these cities, the dataset in \cite{albert_kdd_urbcomp_2018} contains sample spatial maps as square windows of width $W=100km$ around a city center\footnote{The maximum distance from the city center that most people would be willing to commute for work (a one-hour commute driving at $50 \ kmph$) \cite{commute_metropolitan_2016}. Fixing a spatial scale of $W=100km$ results in different image sizes (in pixels) for cities at different latitudes on Earth. For simplicity, we resize all images to $128 \times 128$ pixels.} from four sources:
\par \textit{Built land areas density.} The ``Global Urban Footprint'' (GUF)\cite{esch_sar_guf_2013} is a novel dataset produced by the German Aerospace Center (DLR) that maps the distribution of human settlements for the entire planet at an unprecedented $\sim 12 m/px$ (here we aggregate to $750m/px$). 
\par \textit{Population density data}. The \textit{LandScan} data \cite{landscan-data} consists of population density estimates available worldwide at a $1km/px$ resolution. It has been produced yearly since 2000 at the Oak Ridge National Laboratory using remote-sensing imagery and census surveys. 
\par \textit{Nighttime luminosity data.} NASA's Visible Infrared Imaging Radiometer Suite (VIIRS) \cite{viirs-data} satellite mission provides data on relative luminance values between 20:00 and 22:00 local time at a $750m/px$ resolution and on a scale from 0 (no lights) to 180.
\par \textit{City boundaries.} Data on boundaries at a municipality level was integrated from the Global Administrative Boundaries project (GADM) \cite{gdam-data} (compiled from open source and census surveys).

We further removed those cities for which the amount of signal in the built up layer (fraction of image pixels with non-zero values) was below 2\%, which resulted in a final training dataset of almost $15,000$ world-wide cities. The largest, spatially uniformly-distributed $3,000$ such cities are shown in Figure \ref{fig:3k_cities_map}. 

Several examples of our training set are presented in Figure \ref{fig:example_data}. For each city (column), we present the data sources used, from bottom to top: built areas, luminosity, and population density (the latter two on log-scale). In the left panel, we show several major cities, where the white areas represent the water mask $x_W$ applied to each map layer, and the regions outside official city boundaries ($b(x,y)=0$) are gray-shaded.  

\subsection{Model validation via spatial statistics}

\begin{figure}
    \centering
    \hspace*{-3em}
    \raisebox{-.5\height}{\includegraphics[scale=0.5]{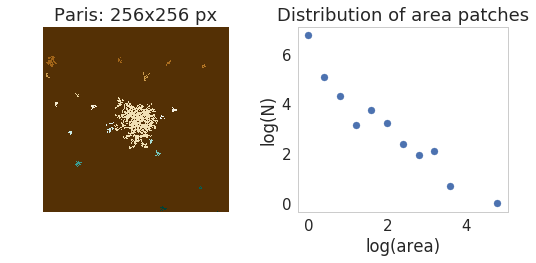}}
    \hspace*{-3em}
    \raisebox{-.4\height}{\includegraphics[scale=0.3]{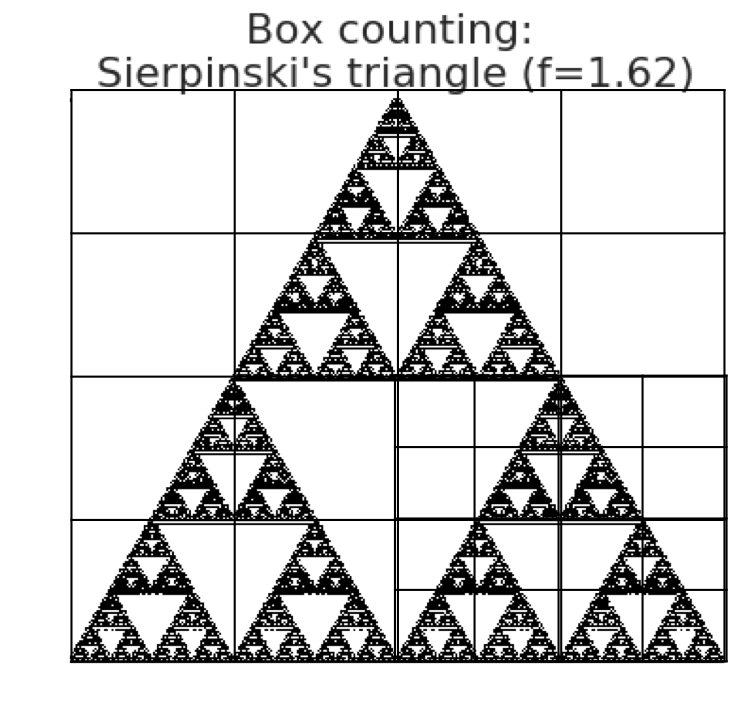}}
    \raisebox{-.45\height}{\includegraphics[scale=0.45]{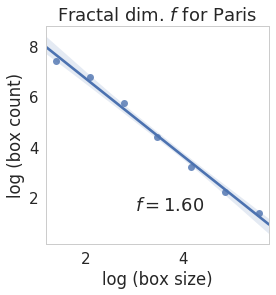}}
    \caption{Spatial statistics for urban form analysis. From left to right: top 20 patches for Paris at a $256px \times 256px$ resolution; log-log distribution of number of built patches with their sizes; illustration of the box counting algorithm \cite{box_counting_paper} for computing fractal dimension $f$; computing $f$ for Paris.}
    \label{fig:built_area_distribution}
\end{figure}

Following typical analysis in statistical geography and urban development \cite{makse1995modelling}, we compute several spatial statistics on the built areas of all our samples, either real or synthetic. The simplest such measure for a given map $x$ is the percentage of built area $a$ in a fixed $W \times W$ window (here $W=200km$) around the city center, $a = \frac{1}{W^2} \sum_{i,j} x_{i,j}$. Next, as a simple extension, we compute the distribution of the sizes of of built agglomerations (contiguous patches) for a given input image $x$\footnote{For this, we use the \texttt{morphology.label} algorithm in \texttt{skimage} package in Python that extracts the connected components of a given image at the pixel level.}. In the top row in Figure \ref{fig:built_area_distribution} we illustrate this computation for Paris. The top-right panel shows the top 20 such contiguous patches, including the large urban core. The top-left panel shows the log-log distribution of the patch size. 

Another commonly-used statistic in ``urban physics" is the \textit{fractal dimension} $f$ of a spatial distribution, which we illustrate in the two rightmost panels in Figure \ref{fig:built_area_distribution}. To compute $f$, we use the classic ``box-counting" algorithm \cite{box_counting_paper}. This algorithm divides up a give input image into a successively finer grid (boxes of sizes $\frac{1}{2}, \frac{1}{4}, \frac{1}{8}, ...$) and calculates the number of boxes at each scale that cover at least a threshold amount of non-zero pixels. This procedure is illustrated in the bottom row in Figure \ref{fig:built_area_distribution} for ``Sierpinski's triangle", a classic fractal shape consisting of self-similar triangles. Then $f$ is computed as the slope of the logarithm of the number of boxes containing enough non-zero pixels with the logarithm of the box size.  

\section{Results}
\label{sec:results}

\subsection{Producing synthetic cities}
\label{sec:results}
\vspace*{-0.1em}

\begin{figure*}[h!]
    \centering
    \includegraphics[scale=0.68]{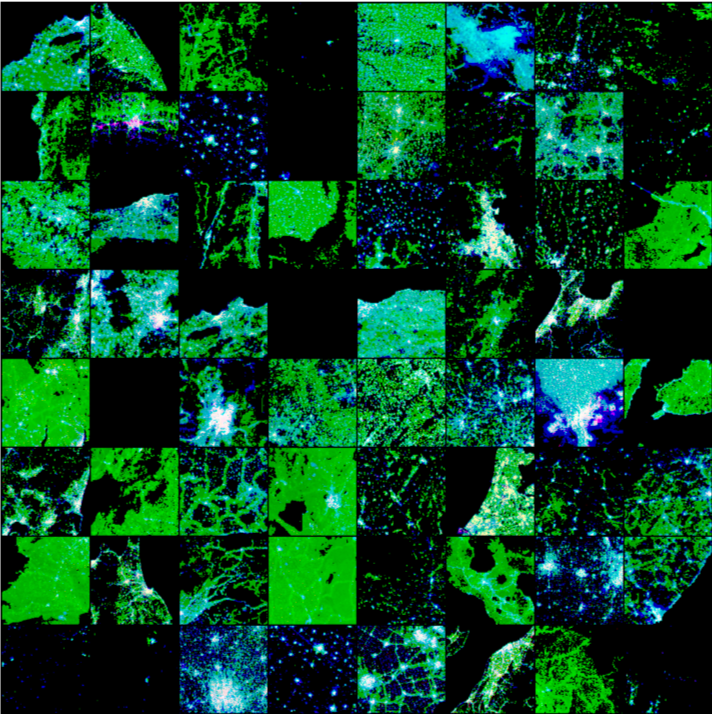}
    \includegraphics[scale=0.66]{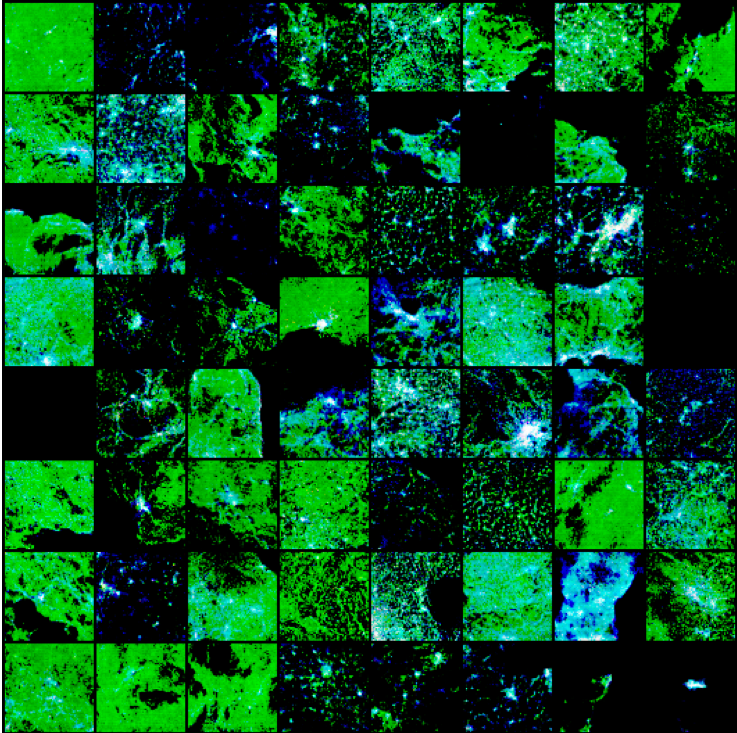}
    \caption{Real (\textit{left}) and synthetic (\textit{right}) cities. For visualization purposes we display population and built areas density maps as green and blue channels. Note the complex, realistic-looking synthetic spatial patterns in the right panel and their qualitative similarity to the real urban patterns on the right.}
    \label{fig:results}
\end{figure*}%

We trained our model from Section \ref{sec:model} using the data from Section \ref{sec:experimental}. We validate the $G$ and $D$ components of the model as follows. 

\par \textbf{Simulating urban forms using $G$.} In Figure \ref{fig:results} (left panel) we show examples of synthetic spatial maps generated by our model that sees only water masks $x_W$ as inputs. To enhance the visualization of the generated patterns, we display population and built areas density maps as green and blue channels. For comparison, we show real, randomly-selected cities on the left panel in the same figure. Note that $G$ is able to generate crisp, realistically-looking urban forms that are virtually undistinguishable from the real cities. 

\begin{figure}[t!]
    \centering
    \raisebox{+.17\height}{\includegraphics[scale=0.4]{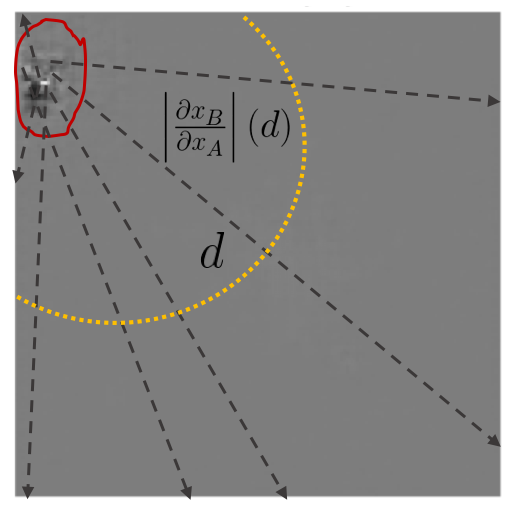}}
    \includegraphics[scale=0.4]{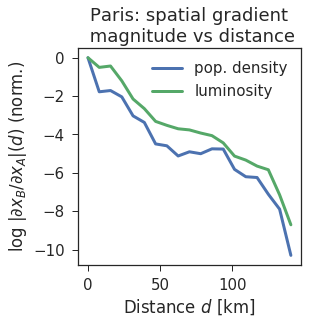}
    \caption{\textit{Left:} sampling rays out of a local region of interest (red outline); \textit{right:} example gradient dependence with distance $d$ from selected region for Paris.}
    \label{fig:stats_comparison}
\end{figure}

Next, we use our GAN-based spatial regression framework to predict built maps $\tilde x_B = G(x_A)$ on the test set. To validate the global structure of the generated built density maps, we compare model-estimated spatial statistics $(\tilde a,\tilde f)$ with ground-truth statistics $(a,f)$ as discussed in Section \ref{sec:experimental}. We show the results of this comparison in Figure \ref{fig:stats_comparison}. Note that the synthetic built density maps are very close to the real built density map, as indicated by values of Pearson's $R^2$ of $\sim0.75$ (for $a$) and $\sim 0.85$ (for $f$). While these domain statistics are relatively simple, more complex ones (e.g., the distribution of the urbanized areas in Figure \ref{fig:built_area_distribution}) can be similarly incorporated. 

\begin{figure}[t!]
    \centering
    \hspace*{-4em}
    \includegraphics[scale=0.4]{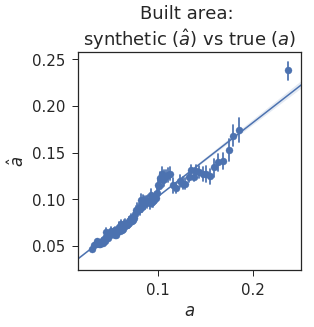}
    \includegraphics[scale=0.4]{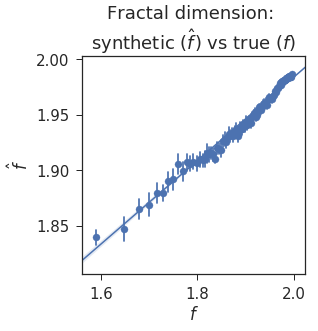}
    \caption{Comparing the ground truth and estimated statistics: the fraction of built area ($a$ vs $\tilde a$, \textit{left}) and the fractal dimension ($f$ vs $\tilde f$, \textit{right}).}
    \vspace*{-1em}    
\end{figure}

\par \textbf{Comparing cities using features learned by $D$.} A second experiment we performed was to extract autoencoder bottleneck layer representations $\phi_i=D^e(\mathbf{x}_i)$ for all cities $i=1,...,N$. As in \cite{urban_environments_paper}, we build a simple classifier (a \texttt{KD-Tree}) that allows to efficiently compare cities by performing nearest-neighbor queries in feature space. This is illustrated in the right panel in Figure \ref{fig:results}, where we show the top $3$ ``most similar'' cities to Paris, San Francisco, Boston, and Lagos. For example, for San Francisco, the model identifies other elongated, coastal towns with complex morphology as top neighbors, whereas for Paris the model returns other circular-shaped cities. This measure of similarity can provide a way to identify ``classes'' of cities by their urban development.

\begin{figure}
    \centering
    \vspace*{-2.2em}    
    \hspace*{-2em}
    \includegraphics[scale=0.36]{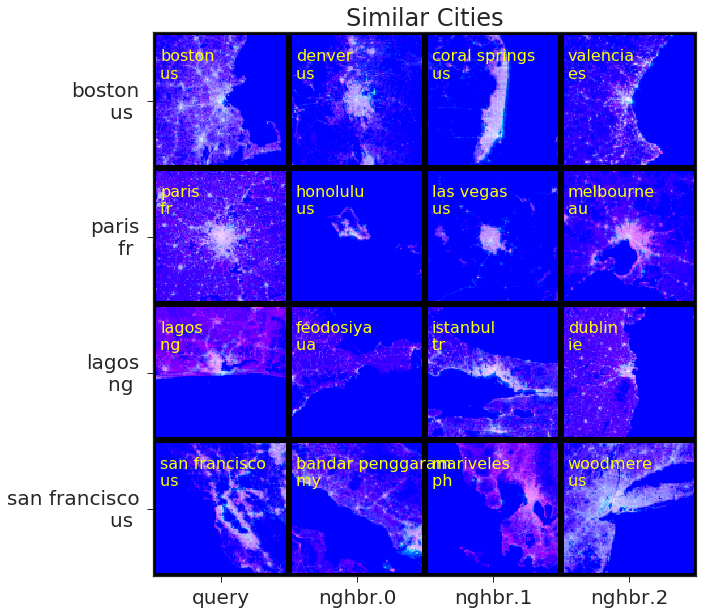}
    \caption{Validating $D$ by using its bottleneck features $\phi$ to assess city similarity. The most similar $3$ cities to each query city are displayed ($x_{bld}, \ x_{pop}, \ x_{lum}$) are stacked for display.}
    \label{fig:results}
\end{figure}%

\subsection{Scenario and sensitivity analysis via GANs}

\begin{figure*}
    \centering
    \hspace*{-3em}
    \raisebox{-0.5\height}{\includegraphics[scale=0.4]{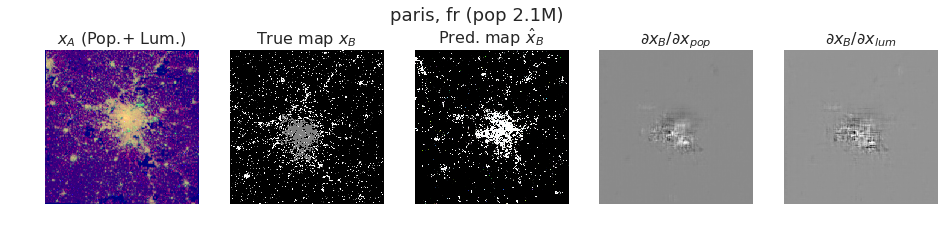}}
    \raisebox{-0.5\height}{\includegraphics[scale=0.3]{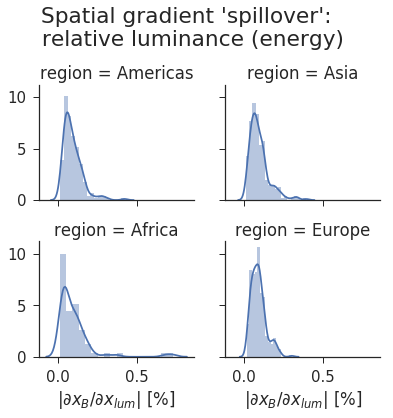}}
    \caption{\textit{Left:} GAN spatial regression example for Paris. The first three columns (from left to right) show the input maps $x_A$, ground truth built areas map $x_B$, model-generated built map $\tilde x_B$. The last two columns show the gradients $\partial x_B/\partial x_\text{pop}$ and $\partial x_B/\partial x_\text{lum}$. \textit{Right}: fraction of gradient outside of a local region of interest $\mathcal R$ produced by a unit change in luminosity in $\mathcal R$.}
    \label{fig:pred_grad_example}
\end{figure*}

We present examples results in Figure \ref{fig:pred_grad_example} for Paris. The first three columns (from left to right) show input population and nightlights map $x_A$ (a multi-channel image), true built map $x_B$, and predicted built map $\tilde x_B=G(x_A)$. The last two panels show the spatial gradient structure for the two components (population density and luminosity). As it is not feasible (or necessary) to compute a full $256 \times 256 \times 2$ gradient matrix for every one of the $256 \times 256$ elements in $x_B$, we compute the spatial gradient for a given local region of interest $\mathcal R$ (highlighted in gray over the true built map $x_B$, second column in the figure) by averaging over all pixels in that region. We simulate two scenarios: \textit{i)} $\mathcal R$ is the main urban core, and \textit{ii)} $\mathcal R$ consists of the top 3 secondary area patches. Note how a ``spillover" effect can be qualitatively observed - in particular for the secondary urbanized areas - where non-zero gradient values appear outside of the areas over which the gradient is computed. This suggests that (static) local changes in the population density (or luminosity) propagate beyond the immediate locality. We next set to characterize the spatial properties of this effect. 

\subsection{Spatial gradient structure and ``spillover"}

Next, we study how a local change in factors related to urban form - population density and economic development (luminosity) - can propagate in space and lead to changes in urban form $\tilde x_B$ potentially relatively far away from where the change originated. Specifically, we compute the dependence of the magnitude of the gradient due to a change in a given local region $\mathcal R$ with the distance from that region. We use the following sampling procedure (see Figure \ref{fig:stats_comparison}):
\begin{enumerate}[leftmargin=*,topsep=0pt,itemsep=-1ex,partopsep=0.5ex,parsep=0.5ex]
    \item For a local region of interest $\mathcal R$ (circled with a red line in the figure, the second-larges urban patch around Paris), generate $m$ (here $m=50$) sampling directions (rays) $r$ uniformly at random;
    \item Compute the magnitude of the gradient $|\partial x_B/ \partial x_A|(d)$ as a function of distance $d$ from the center of $\mathcal R$, by averaging along each $r$ over $\sim 7km$ bins, and then averaging over the $r$s. 
\end{enumerate}

\begin{figure}
    \centering
    \hspace*{-2em}
    \includegraphics[scale=0.33]{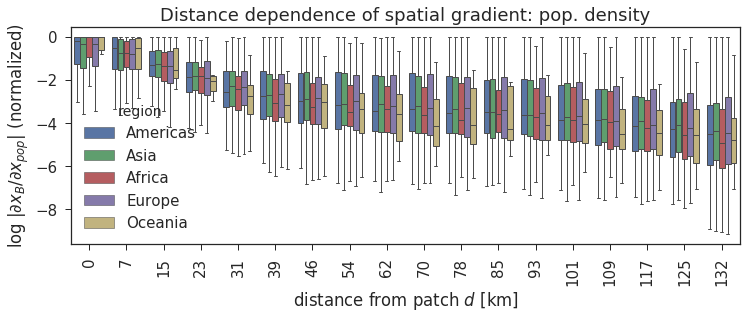}\\
    \caption{Spatial propagation of gradients: (log-normalized) gradient $\partial x_B/\partial x_{pop}$ magnitude with distance from $\mathcal R$.}
    \label{fig:grad_distance_distr}
\end{figure}

An example scenario of applying this sampling method is shown in the rightmost panel in Figure \ref{fig:stats_comparison}. There, we characterize the ``action at a distance" (spatial dependence of the gradient magnitude) of a unit change in either population density (blue curve) or luminosity (green curve) in the second-largest urban region around Paris. Again, we observe that the ``effects" of changes in these factors in the distribution of built environment propagate spatially outward. Next, to quantify this effect at a global level, we apply the same sampling technique over the computed gradient maps with respect to local changes in luminosity and population density over the test dataset of $\sim 3000$ largest metropolitan regions worldwide. This analysis is shown in Figure \ref{fig:grad_distance_distr}. First, we compute the percentage of gradient magnitude outside of the originating region (here we used the top 3 secondary contiguous urban regions outside of the urban core) - what we term ``gradient spillover". The left panel in the figure shows this calculation for cities in four major geographical regions in the case of luminosity. The distributions indicate that spatial gradient ``spillover", while relatively small, is still present. 

Finally, in the right panel in Figure \ref{fig:grad_distance_distr} we present the distance dependence of the average (log-scale, normalized) gradient magnitude, again in the same scenario where a change occurs in the top 3 secondary urban regions outside the urban core, for the case of population density. For each of the $\sim7.5km$ intervals, we show the distribution over the global city sample as box plots, broken down by major geographical region. While there are visible differences across regions, a trend is apparent in that changes propagate, on average, over more than $15km$ before they contribute less than $1\%$ to the spatial gradient at that distance. 
 
\section{Discussion and conclusions}
\label{sec:conclusions}
\vspace*{-0.2em}
To the best of our knowledge, this is the first study in both the machine learning and in the urban analysis literature on a model of urban forms derived from data at a planetary scale. Prior literature on urban analysis almost entirely focuses on cities in developed countries, due to data availability. However, projections show that most of the urbanization in the next 30 years will take place in developing countries, with more than a doubling of the amount of land used in cities in the absence of data-informed regulatory policies. However, in many such places, socio-demographic and economic data is either extremely limited, unreliably reported, or difficult to collect. Together with the poor predictive performance of traditional urban models, this calls for developing better predictive methods, calibrated using a globally-consistent dataset, and is of interest to international planning agencies, real estate markets, and national regulatory agencies.

A flexible generative model such as a GAN is needed because existing models of urban form are not able to reproduce the richness in the spatial distribution of key macroeconomic indicators (building density, population density, economic activity) observed in real data, and require extensive design and calibration to incorporate heterogeneous data. This paper is the first application of a top-down approach to modeling urban spatial maps, where long-range connections between macroeconomic factors are implicitly modeled via a map-to-map regression, without the need of hand-engineering of features. As such, we can uncover the type of global dependencies that we highlight in our analysis of sensitivity (gradient) profiles in a completely non-parametric way. We see this as key differentiator for the task of scenario analysis, and as first step towards a realistic simulator of urban form.

Lastly, we caution that the opacity of GANs (and deep networks more generally) is a common problem in applying these types of models to real decision-making. In absence of ground truth labels of any kind, interpretation is certainly very hard. Clearly, the simple spatial statistics that we use to validate model output do not take into account all of the richness in spatial maps; however they serve as first proof of concept of a GAN-based methodology. Learning disentangled, interpretable representations is certainly an important step towards interpretability and usability. We recognize that, in contrast with other problems typically studied in machine learning (e.g., image classification), in our application there are no straightforward labels to guide training, or, in fact, to allow for validating algorithm output. In future work, we intend to add the capability of conditioning the generation process on any available ancillary information, e.g., macro-economic and geographical regions and city-level data of economic output, diversity etc.


%
%

\bibliographystyle{ACM-Reference-Format}
\bibliography{refs}

\end{document}